\documentclass[letterpaper]{article} % DO NOT CHANGE THIS
\usepackage[submission]{aaai24}  % DO NOT CHANGE THIS
\usepackage{times}  % DO NOT CHANGE THIS
\usepackage{helvet}  % DO NOT CHANGE THIS
\usepackage{courier}  % DO NOT CHANGE THIS
\usepackage[hyphens]{url}  % DO NOT CHANGE THIS
\usepackage{graphicx} % DO NOT CHANGE THIS
\usepackage{natbib}  % DO NOT CHANGE THIS AND DO NOT ADD ANY OPTIONS TO IT
\usepackage{caption} % DO NOT CHANGE THIS AND DO NOT ADD ANY OPTIONS TO IT
\usepackage{algorithm}
\usepackage{algorithmic}
\usepackage{amsfonts}
\usepackage{stmaryrd}
\usepackage{mathrsfs}
\usepackage{amsthm}
\usepackage{amsmath}
\usepackage{booktabs}
\usepackage{newfloat}
\usepackage{listings}
\DeclareCaptionStyle{ruled}{labelfont=normalfont,labelsep=colon,strut=off} % DO NOT CHANGE THIS
\floatstyle{ruled}
\newfloat{listing}{tb}{lst}{}
\floatname{listing}{Listing}
\title{\textit{PhotoVerse}: Tuning-Free Image Customization with Text-to-Image \\ Diffusion Models }
\author{
    Li Chen\textsuperscript{\rm 1}\equalcontrib,
    Mengyi Zhao\textsuperscript{\rm 1, }\textsuperscript{\rm 2}\equalcontrib,
    Yiheng Liu\textsuperscript{\rm 1}\equalcontrib,
    Mingxu Ding\textsuperscript{\rm 1},\\
    Yangyang Song\textsuperscript{\rm 1},
    Shizun Wang\textsuperscript{\rm 1, }\textsuperscript{\rm 3},
    Xu Wang\textsuperscript{\rm 1},
    Hao Yang\textsuperscript{\rm 1},
    Jing Liu\textsuperscript{\rm 1},
    Kang Du\textsuperscript{\rm 1},
    Min Zheng\textsuperscript{\rm 1}
}
\affiliations{
    \textsuperscript{\rm 1}ByteDance, Beijing, China.\\
    \textsuperscript{\rm 2}Beihang University, Beijing, China.\\
    \textsuperscript{\rm 3}National University of Singapore.\\
    \{chenli.phd, liuyiheng.yolo, dingmingxu.leo, songyangyang.2021, wangxu.ailab, jing.liu,\\ dukang.daniel,
    zhengmin.666\}@bytedance.com, zhaomengyi@buaa.edu.cn, \\ shizun.wang@u.nus.edu, yanghao.alexis@foxmail.com
}

\usepackage{bibentry}
\begin{document}

\maketitle

\begin{abstract}
Personalized text-to-image generation has emerged as a powerful and sought-after tool, empowering users to create customized images based on their specific concepts and prompts. However, existing approaches to personalization encounter multiple challenges, including long tuning times, large storage requirements, the necessity for multiple input images per identity, and limitations in preserving identity and editability. To address these obstacles, we present PhotoVerse, an innovative methodology that incorporates a dual-branch conditioning mechanism in both text and image domains, providing effective control over the image generation process. Furthermore, we introduce facial identity loss as a novel component to enhance the preservation of identity during training. Remarkably, our proposed PhotoVerse eliminates the need for test time tuning and relies solely on a single facial photo of the target identity, significantly reducing the resource cost associated with image generation. After a single training phase, our approach enables generating high-quality images within only a few seconds. Moreover, our method can produce diverse images that encompass various scenes and styles. The extensive evaluation demonstrates the superior performance of our approach, which achieves the dual objectives of preserving identity and facilitating editability. Project page: \url{https://photoverse2d.github.io/}

\end{abstract}

\section{Introduction}

The remarkable advancements in text-to-image models, \emph{e.g.}, Imagen \cite{saharia2022photorealistic}, DALL-E2 \cite{ramesh2022hierarchical}, and Stable Diffusion \cite{sd}, have garnered significant attention for their ability to generate photorealistic images based on natural language prompts. Despite the impressive capability of these models to generate diverse and sophisticated images by leveraging large-scale text-image datasets, they encounter difficulties when it comes to synthesizing desired novel concepts. Consider the scenario where users aim to incorporate themselves, family members, or friends into a new scene, achieving the desired level of fidelity solely through text descriptions becomes challenging. This challenge stems from the fact that these novel concepts were absent from the dataset used for training the models, making it hard to generate accurate representations based solely on textual information.
In the pursuit of enabling personalized text-to-image synthesis, several methods \emph{e.g.}, Dreambooth \cite{ruiz2023dreambooth}, Textual Inversion \cite{TI}, DreamArtist \cite{dong2022dreamartist}, and CustomDiffusion \cite{kumari2023multi} primarily focused on identity preservation and propose the inverse transformation of reference images into the pseudo word through per-subject optimization. Text-to-image models undergo joint fine-tuning to enhance fidelity. The resulting optimized pseudo word can then be leveraged in new prompts to generate scenes incorporating the specified concepts.
\begin{figure}[!t]
	\includegraphics[width=1\linewidth]{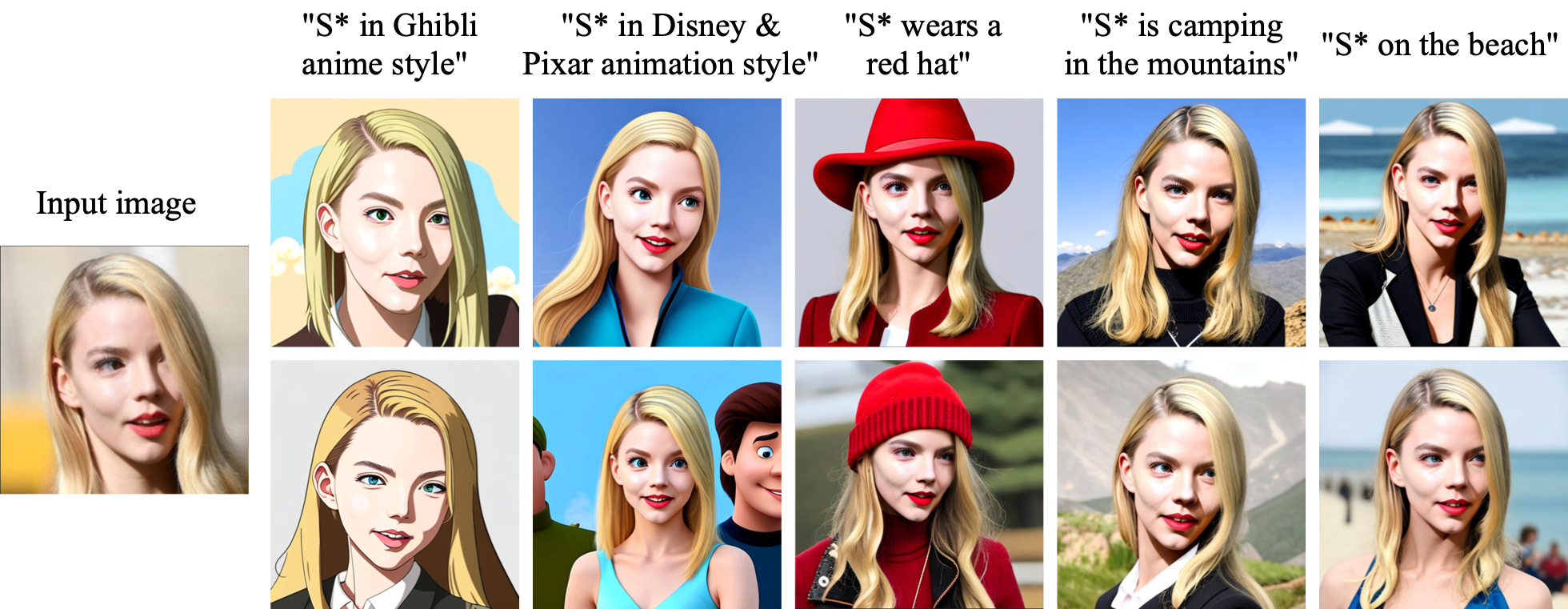}
	\caption{Our proposed PhotoVerse exhibits a wide range of diverse results. By providing a single reference image featuring the target concept alongside various prompts, PhotoVerse facilitates the generation of high-quality images that align seamlessly with the given prompts. Notably, our method achieves this outcome within a matter of seconds, eliminating the need for test-time tuning.}
\end{figure}
However, this optimization-based paradigm often requires expensive computational resources and large storage requirements, taking minutes to hours when executed on high-end GPUs. Such lengthy processing times and tuning processes render it impractical for user-friendly usage, where short time generation is desirable. Also, these approaches had limitations in their language editing capabilities due to potential overfitting on a tiny identity-specific image dataset.
Recent encoder-based methods \emph{e.g.}, E4T \cite{E4T}, InstantBooth \cite{shi2023instantbooth}, SuTI \cite{chen2023subject}, Profusion \cite{zhou2023enhancing} aimed to address these challenges by optimizing pseudo token embeddings, introducing sampling steps, or incorporating new modules for concept injection. However, challenges persist, such as the need for multiple input images and test-time tuning \cite{sohn2023styledrop} to maintain identity preservation and editability.

To address the aforementioned issues, we present a novel methodology that leverages a dual-branch conditioning mechanism, combining improved identity textual embeddings and spatial concept cues through dual-modality adapters in both the text and image domains. Furthermore, we introduce a facial identity loss component during training to enhance identity preservation. Our method stands out by effectively balancing high identity similarity and robust editing capabilities.
Crucially, our approach achieves a remarkable reduction in generation time for personalized text-to-image models, completing the process in just a few seconds and utilizing only a single concept photo. This significant advancement in efficiency streamlines the generation workflow and enhances the user experience.

In summary, our contributions can be categorized into three key aspects:
\begin{itemize}
\item We propose a novel architecture for user-friendly text-to-image personalization. Our approach eliminates the need for test-time tuning and requires only a single image of the target subject. This results in a rapid and effortless generation, typically completed in $\sim$ 5 seconds.

\item We introduce a dual-branch concept injection paradigm that effectively extracts identity information from both textual embeddings and visual representations. By leveraging this approach, we enhance the preservation of identity during training. Additionally, we incorporate a face identity loss component to further facilitate identity preservation throughout the training process.

\item We demonstrate the exceptional quality of our method in maintaining identity while capturing rich details and essential attributes, such as facial features, expressions, hair color and hairstyle. Our approach not only ensures identity preservation but also preserves editability. It empowers diverse stylization, image editing, and new scene generation in accordance with the provided prompts.
\end{itemize}

\section{Related Work}

\subsubsection{Text-to-Image Synthesis}
The text-to-image synthesis relies on deep generative models such as Generative Adversarial Networks (GANs) \cite{xia2021tedigan, kang2023scaling}, autoregressive models \cite{ramesh2021zero}, and diffusion models \cite{ho2020denoising, sd}. Initially, prior works primarily focused on generating images under specific domain and text conditions, limiting their applicability. However, with the advent of large-scale image-text datasets and powerful language encoders, text-to-image synthesis has achieved remarkable advancements. DALL-E \cite{ramesh2021zero} was the first approach utilizing an autoregressive model to generate diverse and intricate images from arbitrary natural language descriptions. This methodology served as the foundation for subsequent methods like Make-A-Scene \cite{gafni2022make}, CogView \cite{ding2021cogview}, Parti \cite{yu2022scaling}, Muse \cite{chang2023muse}, and others. 

However, the pioneering work of GLIDE \cite{nichol2021glide} introduced diffusion models, surpassing the autoregressive-based DALL-E in producing more photorealistic and high-resolution images. Consequently, diffusion models have gradually emerged as the predominant approach for text-to-image synthesis. Subsequent advancements, such as DALL-E2 \cite{ramesh2022hierarchical}, Imagen \cite{saharia2022photorealistic}, and LDM \cite{sd}, have further improved diffusion models in terms of photorealism, language comprehension, and generation diversity. Notably, the release of Stable Diffusion as an open-source model has propelled its widespread adoption, making it the most extensively utilized text-to-image model. This has also led to the creation of numerous fine-tuned models by the research community. Given this context, we employ Stable Diffusion in our experiments.

\subsubsection{Image Inversion}
In the domain of generative models, the ability to invert an image into a latent code that accurately reconstructs the original image holds significant importance. This capability facilitates various downstream applications that rely on manipulating the latent code to enable tasks such as image editing, translation, customization, and overall control over image generation. In the literature on GAN inversion, there are primarily two approaches: optimization-based inversion, involves directly optimizing the latent code to minimize image reconstruction error. While this method achieves high fidelity, its drawback lies in the requirement of a hundred times iterations, rendering it unsuitable for real-time applications. Encoder-based inversion, trains a neural network to predict the latent code. Once trained, the encoder performs a single forward pass to obtain a generalized result, offering improved efficiency.

The inversion of diffusion models also draws inspiration from the aforementioned methods. However, due to the iterative nature of diffusion model-based image generation, inverting an image into the noise space associated with the image (\emph{e.g.}, DDIM \cite{song2020denoising}, DDIB \cite{su2022dual}, ILVR \cite{choi2021ilvr}, CycleDiffusion \cite{wu2022unifying}) results in a latent space that is not as decoupled and smooth as the latent space of GANs. Consequently, identity preservation suffers. Alternatively, some works explore performing inversion in a different latent space, such as textual embedding space \cite{TI}. This space exhibits strong semantic information and better preserves the object's characteristics. In our research, we employ an encoder-based approach to achieve instant inversion in the text embedding space. And the method is further extended to conditioning on visual features, which can quickly capture the image in multi-modality, and realize fast generation.

\subsubsection{Personalization}

By leveraging the generative capabilities of pre-trained text-to-image models, Personalization offers users the ability to synthesize specific unseen concepts within new scenes using reference images. Typically, the unseen concept is transformed into a pseudo word (\emph{e.g.}, $S^{*}$) within the textual embedding space \cite{ruiz2023dreambooth, TI}. Optimization-based methods, such as DreamArtist \cite{dong2022dreamartist}, directly optimize the pseudo word to establish a mapping between user-provided images and textual inversion. Other approaches, \emph{e.g.}, Dreambooth \cite{ruiz2023dreambooth} and CustomDiffusion \cite{kumari2023multi} employ fine-tuning of text-to-image models to enhance fidelity. However, these strategies require minutes to hours for concept-specific optimization. In contrast, encoder-based methods such as E4T \cite{E4T}, InstantBooth \cite{shi2023instantbooth}, Profusion \cite{zhou2023enhancing} train an encoder to predict the pseudo word, enabling the generation of personalized images within a few fine-tuning steps. 

Nonetheless, tuning the entire model entails substantial storage costs and often results in overfitting on specific concepts. Moreover, many approaches \cite{TI, ruiz2023dreambooth,kumari2023multi, shi2023instantbooth,chen2023subject} rely on multiple reference images, which is not always the case users could provide. To address these limitations, our work mitigates these imperfections through the utilization of the parameter-efficient fine-tuning technique and the design of an encoder that can perform the personalization task with only one reference image, enhancing the efficiency and effectiveness of the synthesis process.

%todo add 需要 test time finetune的方法  SuTI \cite{chen2023subject},
\section{Methodology}

The objective of personalized text-to-image synthesis is to train models to learn specific concepts through reference images and subsequently generate new images based on text prompts. In our paper, we aim to achieve instantaneous and optimization-free personalization using only a single reference image. To accomplish this, we propose a novel approach for image customization by integrating a dual-branch conditioning mechanism in both the textual and visual domains. This involves designing adapters to project the reference image into a pseudo word and image feature that accurately represents the concept. These concepts are then injected into the text-to-image model to enhance the fidelity of the generated personalized appearance.

\begin{figure}[t]
\centering
\includegraphics[width=0.9\linewidth]{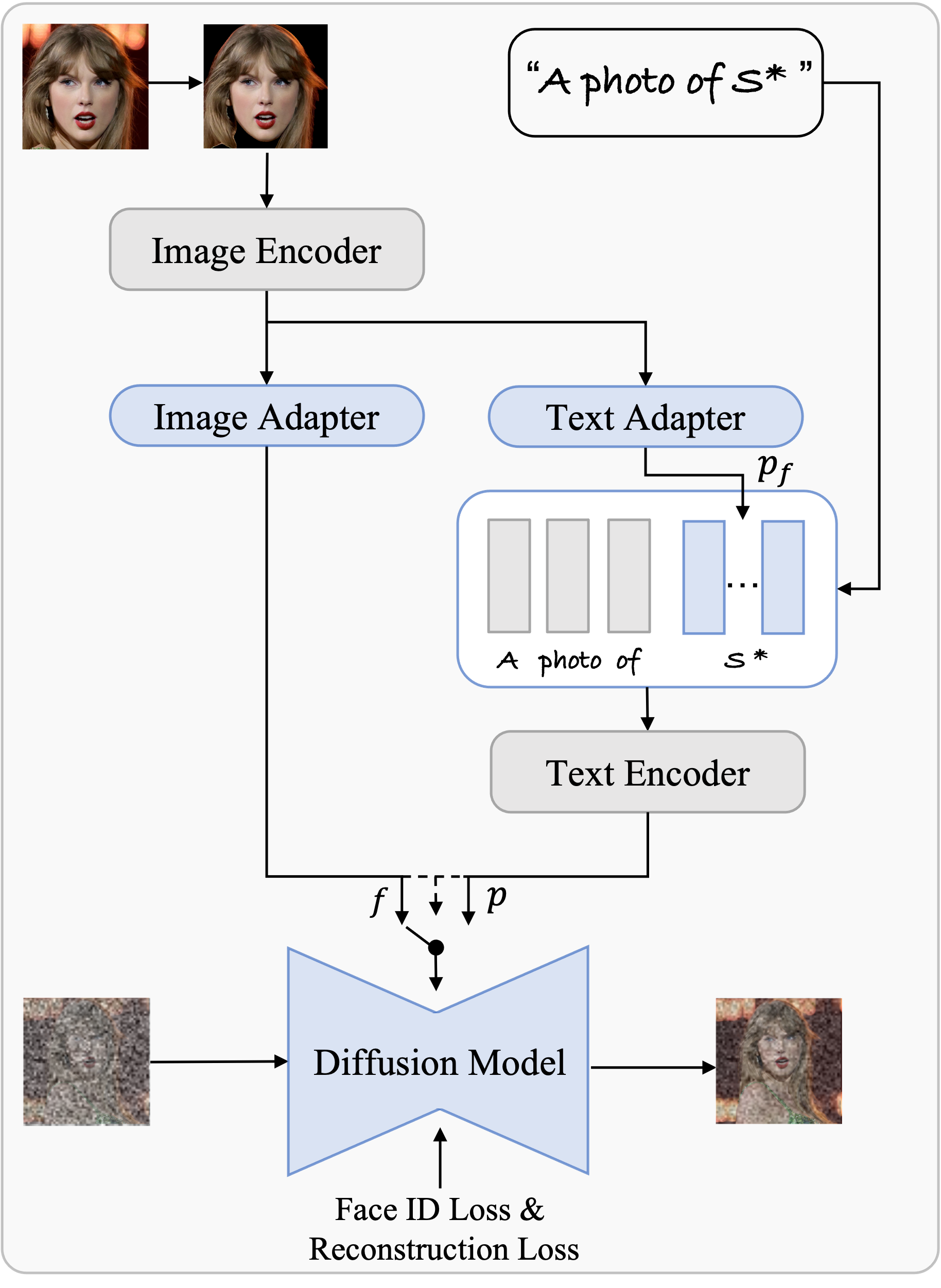} 
	\caption{Overview of our proposed PhotoVerse.}
\label{fig:framework} 
\end{figure}

To enable this process, we incorporate the original text-to-image model with concept conditions and train it within a concept scope, supplemented by an additional face identity loss. A high-level overview of our proposed method is depicted in Figure \ref{fig:framework}.

\subsection{Preliminary}

We utilize Stable Diffusion \cite{sd} as our base text-to-image model, which has demonstrated stability and reliability in the field. This model is trained on a large-scale dataset of image-text pairs, enabling it to generate high-quality and diverse images that align with the given text prompts. Stable Diffusion, based on the Latent Diffusion Model (LDM) architecture, comprises two crucial components.

Firstly, an autoencoder $(\mathcal{E}, \mathcal{D})$ is trained on a large-scale image dataset to compress images. The encoder $\mathcal{E}$ maps an image $x$ from the pixel space to a low-dimensional latent space $z=\mathcal{E}(x)$. The decoder $\mathcal{D}$ is responsible for reconstructing the latent representation $z$ back into an image with high fidelity, aiming to achieve $\mathcal{D}(\mathcal{E}(x))\approx x$.

Secondly, a denoising network $\mathbb{E}$, utilizing the UNet \cite{unet} architecture as the backbone, is trained to perform the diffusion process in the latent space. This approach offers computational and memory advantages compared to operating directly in the pixel space. Additionally, the diffusion model incorporates a text condition $y$ for text-to-image synthesis. It employs a CLIP text encoder $c_\theta$ to project the condition $y$ into an intermediate representation $c_\theta(y)$. This representation is then employed in the intermediate layers of the UNet through a cross-attention mechanism:
\begin{equation}
Attn(Q,K,V)=\text{Softmax}(\frac{QK^T}{\sqrt{d'}})V,
\end{equation}
where $Q=W_Q \cdot \varphi\left(z_t\right), K=W_K \cdot c_\theta(y), V=W_V \cdot c_\theta(y)$, $\varphi\left(z_t\right)$ is the hidden states inside Unet, $z_t$ is the latent noised to time $t$, $d'$ corresponds to the scale factor utilized for attention mechanisms. The training objective of the latent diffusion model is to predict the noise that is added to the latent of the image, which is formulated as:
\begin{equation}
\mathcal{L}_{\text{diffusion}}=\mathbb{E}_{z \sim \mathcal{E}(x), y, \epsilon \sim \mathcal{N}(0,I), t}\left[\left\|\epsilon-\epsilon_\theta\left(z_t, t, c_\theta(y)\right)\right\|_2^2\right],
\end{equation}
here $\epsilon$ represents the unscaled noise sample, and $\mathbb{E}$ denotes the denoising network. During inference, a random Gaussian noise $z_T$ is sampled, and through a series of $T$ iterative denoising steps, it is transformed into $z_0'$. Subsequently, the decoder $\mathcal{D}$ is employed to generate the final image, given by $x'=\mathcal{D}(z_0')$.

\subsection{Dual-branch Concept Extraction}

Prior to extracting facial features, it is essential to preprocess the input images. In the preprocessing stage, Firstly, a face detection algorithm was applied to identify and localize human faces within the input image $x$. This step ensured that the subsequent analysis focused solely on facial regions of interest. Additionally, to provide a slightly larger region for feature extraction, the bounding boxes around the detected faces were expanded by a scaling factor \emph{i.e.}, 1.3. This expansion ensured that important facial details were captured within the region of interest. Subsequently, the expanded regions are cropped as a new face image $x_m$ for subsequent feature extraction. Furthermore, to meet the specific image generation requirements of the diffusion model, the expanded facial regions were resized to the desired shape. 
Moreover, to remove non-facial elements and focus solely on the facial attributes, a mask was applied to the resized facial regions. This mask effectively masked out irrelevant areas such as background and accessories, ensuring that subsequent identity feature extraction conduct efficiently. For the denoising image during training of the Diffusion model, we also employ the cropped face image with a scaling factor of 3 as $x_t$.

\noindent \textbf{Textual Condition.} To personalize a specific concept that cannot be adequately described by existing words, we adopt the approach proposed in \cite{TI, elite} to embed the reference image into the textual word embedding space. In this process, we summarize the concept using pseudo-words denoted as $S^*$. Initially, we utilize the CLIP image encoder, the same encoder employed in Stable Diffusion, to extract the features of the reference image. Following the technique described in \cite{elite}, we enhance the representational capability of the image tokens as well as the editability of the model by selecting features after $m$ layers from the CLIP image encoder, which capture spatial and semantic information at various hierarchical levels of abstraction and concreteness.
Subsequently, a multi-adapter architecture is employed to translate the image features from each layer into multi-word embeddings, resulting in $S^*={{S_1},...,{S_m}}$. Since CLIP effectively aligns textual embeddings and image features, each text adapter consists of only two MLP layers with non-linear activations, making it lightweight and enabling fast representation translation. This design choice leverages the existing alignment provided by CLIP, ensuring efficient and accurate transformation of image features into textual embeddings.

\noindent \textbf{Visual Condition.} 
Despite the advantages of condition in the textual embedding space, there are certain limitations to consider. For instance, the performance can be influenced by the encoder ability of the following text encoder, and the semantic information in the text space tends to be abstract which leads to higher requirements for token representation capabilities. Consequently, we propose the incorporation of the condition from the image space as an auxiliary aid, which is more specific for the model to understand new concepts and contributes to the effective learning of pseudo-tokens in the text space. 
To accomplish this, we utilize the feature obtained from the CLIP image encoder. These features are then mapped using an image adapter, which follows the same structural design as the text adapter. The resulting feature capture essential visual cues related to identity, enabling a more accurate representation of the desired attributes during image generation.

\subsection{Dual-branch Concept Injection}

\begin{figure}[t]
\centering
	\includegraphics[width=0.9\linewidth]{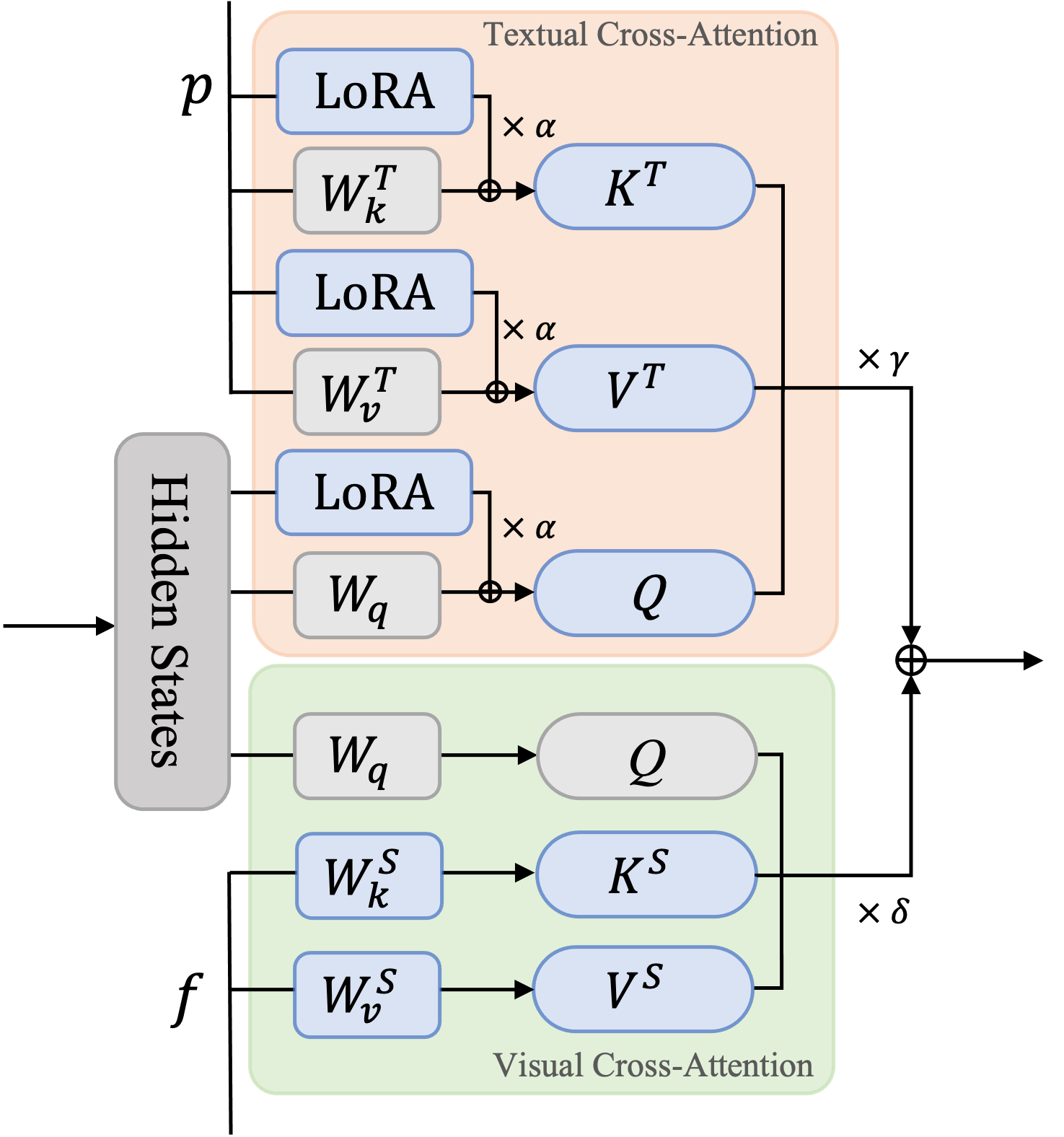} 
	\caption{Concept injection mechanism in the cross-attention module of UNet.}
\label{fig:conjection} 
\end{figure}

The original diffusion model might lack the necessary features required for constructing novel concepts. Thus, we propose injecting the new concept into the text-to-image synthesis process by fine-tuning the diffusion model to recover personal appearance concepts with a high level of detail.
Rather than fine-tuning the entire UNet, which can be computationally expensive and potentially reduce model editability due to overfitting, we only add conditions and fine-tune the weights in the cross-attention module. Previous studies, such as E4T \cite{E4T} and Custom Diffusion \cite{kumari2023multi} have also highlighted that the most expressive model parameters reside in attention layers.

As shown in Figure \ref{fig:conjection}, for the textual conditioning branch, we incorporate the Parameter-Efficient Fine-Tuning (PEFT) method \emph{i.e.}, LoRA \cite{lora}, a more efficient tuning approach. LoRA was originally designed to adapt large-scale pre-trained natural language models to new downstream tasks or domains while preserving performance. LoRA works by freezing the pre-trained model weights and introducing new trainable rank decomposition matrices into each layer of the model architecture. Consequently, we have $K^T = {W_k}^T p + \alpha \Delta {W_k}^T p$ and $V^T = {W_v}^T p + \alpha \Delta {W_v}^T p$, where $p$ represents the text feature extracted from the text encoder, $\Delta {W_k}^T$ and $\Delta {W_v}^T$ denote the low-rank decomposition $\Delta {W} = BA$, with $A$ and $B$ containing trainable parameters.
Regarding the visual conditioning branch, $K^S ={W_k}^S f$ and $V^S = {W_v}^S f$, where $f$ corresponds to the image representation obtained after passing through the image adapter.

Then we present a random fusion strategy for the multi-modality branches:
\begin{equation}
    O = \gamma Attn(Q,K^T,V^T) + \sigma Attn(Q,K^S,V^S),
\end{equation}
where $\gamma$ and $\sigma$ denote two scale factors that regulate the influence of control.
A random seed is sampled from the uniform distribution $\mathcal{U}=(0,1)$, the fused representation can be obtained in the following manner:
\begin{equation}
O=\left\{\begin{array}{ll}
\gamma Attn(Q,K^T,V^T), & \text { if }seed < r_1; \gamma=2 \\
\sigma Attn(Q,K^S,V^S), & \text { if }seed > r_2; \sigma=2 \\
\gamma Attn(Q,K^T,V^T) + \\ \sigma Attn(Q,K^S,V^S), & \text { otherwise }; \gamma=\sigma=1
\end{array}\right.
\end{equation}
where $r_1$ and $r_2$ is the threshold of random $seed$.

In addition, we introduce regularization for both persuade token embedding after text adapter $p_f$ and reference image values $V^S$. Specifically, 
\begin{equation}
{\mathcal{L}^T_{reg}}=  Mean\|p_f\|_1,
\end{equation}
and 
\begin{equation}
{\mathcal{L}^S_{reg}}= Mean\|V^S\|_1.
\end{equation}

The whole pipeline can be trained as usual LDM \cite{sd} does, except for additional facial identity preserving loss $L_{face}$:
\begin{equation}
\mathcal{L}_{face}=\mathcal{C}(f(x)-f(x')).
\end{equation}
Here, $f(\cdot)$ represents a domain-specific feature extractor, $x$ denotes the reference image, and $x'$ corresponds to the denoised image with the prompt ``a photo of $S^*$". To achieve the goal of measuring identity in human face scenarios, we employ the Arcface face recognition approach \cite{deng2019arcface}. The function $\mathcal{C}(\cdot)$ computes the cosine similarity of features extracted from the face region in the images.
Subsequently, the total training loss is formulated as:
\begin{equation}
\mathcal{L}_{total}=\mathcal{L}_{\text{diffusion}}+\lambda_{face} \mathcal{L}_{face} + \lambda_{rt}\mathcal{L}^T_{reg} + \lambda_{rv} \mathcal{L}^S_{reg},
\end{equation}
where $\lambda_{face}$ is the scale factor of $\mathcal{L}_{face}$, while $\lambda{rt}$ and $\lambda_{rv}$ are hyperparameters that control the sparsity level for the specific feature.

Overall, the lightweight adapters and UNet are jointly trained on public datasets within a concept scope, such as \cite{ffhq}. In this way, PhotoVerse can learn how to recognize novel concepts. At inference time, fine-tuning the model is obviated, thereby enabling fast personalization based on user-provided textual prompts.

\begin{figure*}[!t]
\centering
	\includegraphics[width=0.9\linewidth]{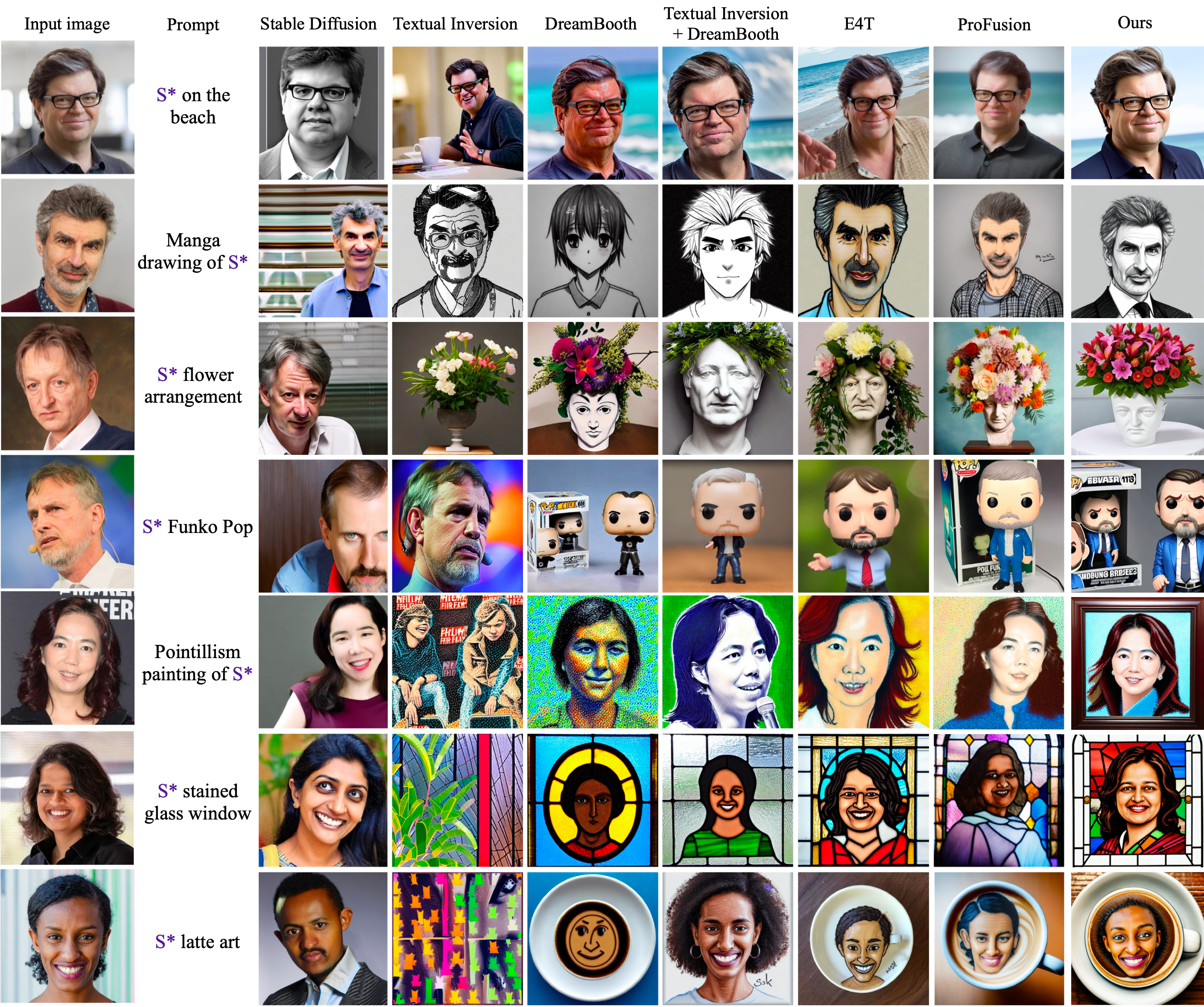}
	\caption{Comparison with State-of-the-Art methods. Our proposed PhotoVerse shows superior performance in preserving identity attributes and generating high-quality images. Notice that DreamBooth, Textual Inversion, E4T and ProFusion require an additional stage of fine-tuning on the provided reference image prior to generation. In contrast, our PhotoVerse is tuning-free and boasts rapid generation speed, offering a distinct advantage in terms of efficiency and user convenience.}
\label{fig:comparison}
\end{figure*}
\section{Experiments}
\subsection{Experimental settings}
\noindent \textbf{Datasets.} We fine-tune our model on three public datasets Fairface \cite{karkkainen2019fairface}, CelebA-HQ \cite{celea-hq} and FFHQ \cite{ffhq}. Evaluation is conducted on a self-collected dataset, which contains $326$ images with balanced racial and gender samples for 4 colors: White, Black, Brown, and Yellow from all the race groups.

For the evaluation phase, we generate five images for each photo with the default inversion prompt ``a photo of $S^*$". This allows us to robustly measure the performances and generalization capabilities of our method.

\noindent \textbf{Implementation Details.} For the text-to-image diffusion model, we utilize Stable Diffusion \cite{sd} as the base model and fine-tune adapters, loRA as well as visual attention weights. We pre-train the model with learning rate of $1e-4$ and batch size of $64$ on V100 GPU. Our model was trained for $60,000$ steps. During training, $\alpha=1$, $\lambda_{face}=0.01$, $\lambda_{rt}=0.01$, $\lambda_{rv}=0.001$, $m=5$, $r_1=\frac{1}{3}$, $r_2=\frac{2}{3}$. For evaluation, we set $m=1$, $\alpha=1$ during sampling. In addition, one can adjust parameters $m$, $\sigma$, $\gamma$, and $\alpha$ with more flexibility to achieve desired generation objectives, such as preserving more identity for photo editing or incorporating stronger style effects for the Ghibli anime style. By customizing these parameters, one can set the level of ID information retention and the degree of stylization according to specific preferences and application requirements. Besides, one can also integrate the image-to-image trick during sampling to reduce randomness, which means replacing initial noise $z_T$ with $z_t$ and sampling from time step $t$, here $z_t$ is derived from the noisy image of $x_0$.

\noindent \textbf{Evaluation Metrics.} In this paper, we aim to assess the ability of generated results to preserve identity by employing facial ID similarity as an evaluation metric. Specifically, we utilize a facial recognition model \emph{i.e.}, Arcface \cite{deng2019arcface} to compute the cosine similarity between facial features. The ID similarity serves as a quantitative measure to determine the extent to which the generated outputs retain the original identity information. 

\subsection{Quantitative Results}

\begin{figure}[!ht]
	\includegraphics[width=1\linewidth]{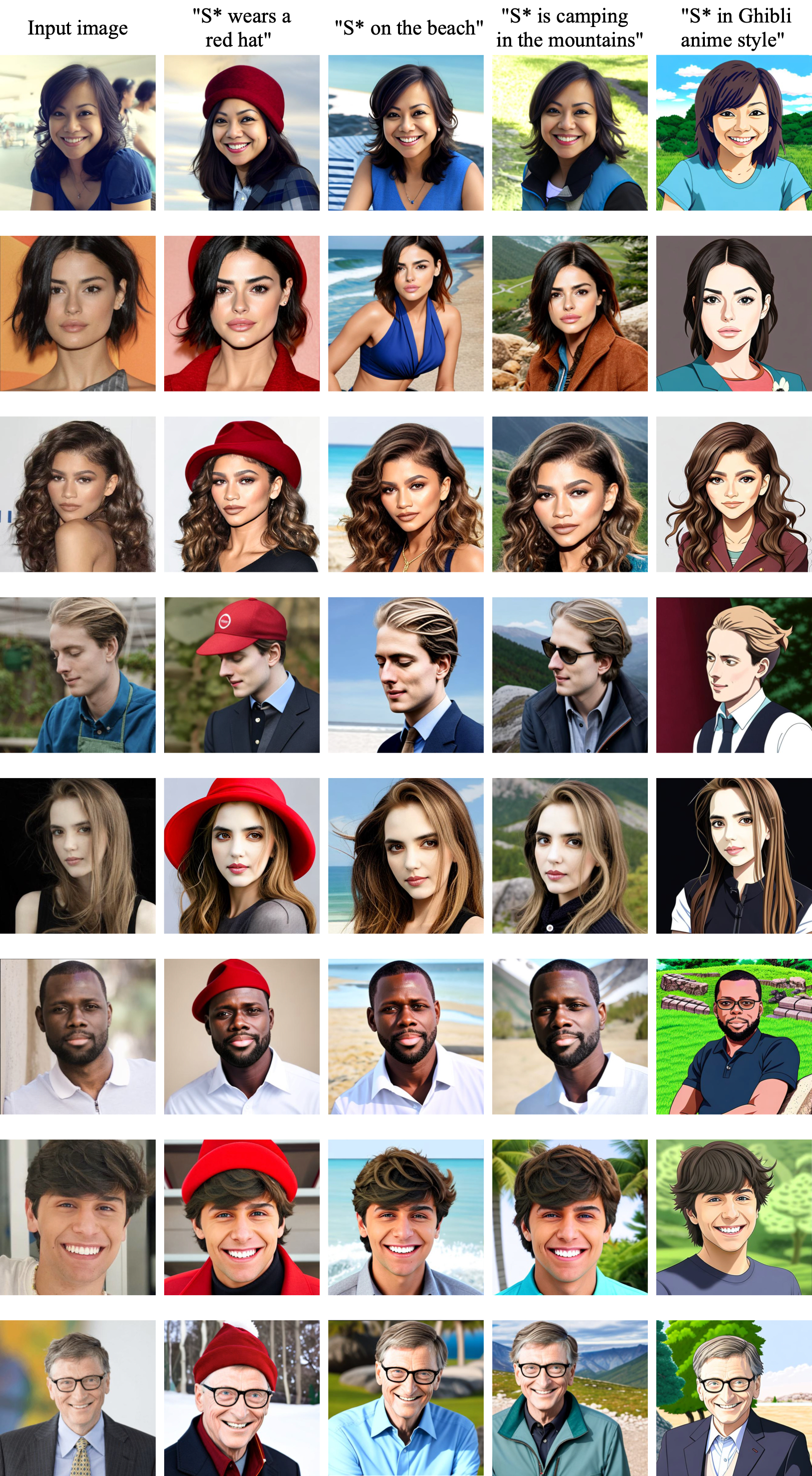}
	\caption{Results of our PhotoVerse with various prompts for stylization and new scene generation.}
\label{fig:results}
\end{figure}
As indicated in Table \ref{table:ablation}, we conducted an evaluation to assess the facial identity similarity in generated facial images. The evaluation involved measuring the ID similarity across diverse ethnicities by employing cosine similarity on facial features. The obtained results substantiate the efficacy of our approach, which achieves a notable level of similarity, with an average value of $0.696$. Remarkably, certain ethnicities, including brown and white, exhibit similarity scores surpassing $0.7$, with white ethnicity demonstrating the highest similarity. We posit that this discrepancy could be attributed to the inherent bias of the pre-trained large model. Regardless, it is noteworthy that our method demonstrates consistent and robust performance across all ethnicities, with a marginal deviation of approximately $0.03$ observed among different ethnic groups.

\subsection{Qualitative Results}
 
Within Figure \ref{fig:comparison}, we present a comprehensive comparison of our proposed PhotoVerse method alongside state-of-the-art (SOTA) personalization techniques, namely DreamBooth \cite{ruiz2023dreambooth}, Textual Inversion \cite{TI}, E4T \cite{gal2023encoder}, and ProFusion \cite{zhou2023enhancing}, focusing on qualitative results. The results of corresponding methods are directly taken from \cite{zhou2023enhancing}. Notably, all aforementioned methods require test time tuning. For instance, DreamBooth and Textual Inversion necessitate 3-5 photos per subject, which incurs considerable time and storage requirements for personalization. DreamBooth, in particular, demands approximately 5 minutes, while Textual Inversion entails 5000 steps for fine-tuning. E4T and ProFusion allow customization with just 1 reference image, but they still require additional time for fine-tuning, approximately 30 seconds \cite{zhou2023enhancing}. In contrast, our proposed approach is test time tuning-free, enabling the synthesis of 5 images within a mere 25 seconds. This remarkable efficiency makes our method exceedingly user-friendly, significantly enhancing the user experience.

Furthermore, concerning the preservation of identity attributes in the reference image, our PhotoVerse (shown in the last column) exhibits exceptional proficiency in capturing facial identity information. Our approach successfully retains crucial subject features, including facial features, expressions, hair color, and hairstyle. For instance, when compared to alternative techniques, our proposed method outperforms in restoring intricate hair details while effectively preserving facial features, as evident in the first row. Additionally, as observed in the second row, our approach excels at fully restoring facial features while maintaining consistency with the desired ``Manga" style specified in the prompt. In contrast, the Profusion-generated photo exhibits blurry mouth details, while E4T fails to exhibit a sufficiently pronounced ``Manga" style. Shifting to the third row, our results successfully capture characteristic expressions present in the input images, such as frowning.

Regarding the quality of generated images, our method surpasses other works by producing noticeably sharper images with enhanced detail. Moreover, our approach exhibits a high degree of aesthetic appeal, featuring natural lighting, natural colors, and the absence of artifacts.

Figure \ref{fig:results} presents additional results showcasing the performance of our approach across different ethnicities in image editing, image stylization, and novel scene generation, further substantiating the three aforementioned aspects of superiority.

\subsection{Ablation studies}
\noindent \textbf{Effect of Visual conditioning branch} According to Table \ref{table:ablation}, it is evident that the image branch has a substantial impact on the preservation of ID information. Removing the image branch leads to a loss of $0.14$, indicating its crucial role in the model's ability to maintain identity consistency. This effect can be attributed to the provision of specific, detailed, and spatial conditions by the visual branch during image generation. Moreover, the visual branch contributes positively to optimizing the embedding of textual tokens.

\noindent \textbf{Effect of Regularizations} The experimental results illustrate the importance of regularizations for visual values and textual facial embeddings during concept injection. It can promote the sparsity of representations, thereby retaining key values and mitigating overfitting issues, as well as enhancing the generalization capability of our model.

\noindent \textbf{Effect of Face ID loss}
We also evaluated the impact of face ID loss on the preservation of identity information. The experimental results demonstrate that it also plays an important role in maintaining identity, resulting in an improvement of $0.05$ in performance.

\begin{table}[!t]
\resizebox{0.485\textwidth}{!}{
\begin{tabular}{l|lllll}
\bottomrule
Method & Black & Brown & White & Yellow & All   \\ \hline
w/o visual conditioning branch  & 0.561 & 0.563 & 0.584 & 0.556  & 0.556 \\
w/o $\mathcal{L}^S_{reg}$  &  0.566 & 0.573  &   0.589 &  0.550 &  0.569   \\
w/o $\mathcal{L}_{face}$    & 0.632 & 0.658 & 0.663 & 0.622  & 0.643 \\ 
w/o $\mathcal{L}^T_{reg} $    &  0.650 & 0.668 & 0.678 & 0.657 &       0.663 \\\hline
PhotoVerse & \textbf{0.685} & \textbf{0.702} & \textbf{0.715} & \textbf{0.682}  & \textbf{0.696} \\ \bottomrule
\end{tabular}}
\caption{Ablation study results of 4 components on ID similarity of 4 races.}
\label{table:ablation}

\end{table}

\section{Conclusions}
In this paper, we introduced an innovative methodology that incorporates a dual-branch conditioning mechanism in both the text and image domains. This approach provides effective control over the image generation process. Additionally, we introduced facial identity loss as a novel component to enhance identity preservation during training.
Remarkably, our proposed PhotoVerse eliminates the need for test-time tuning and relies solely on a single facial photo of the target identity. This significantly reduces the resource costs associated with image generation. Following a single training phase, our approach enables the generation of high-quality images within just a few seconds. Moreover, our method excels in producing diverse images encompassing various scenes and styles. Our approach also supports incorporating other methods such as ControlNet \cite{zhang2023adding}, specifically leveraging its control branch for preserving the overall high-level structure, further enhancing the pose control of personalized text-to-image generation. 
Through extensive evaluation, we have demonstrated the superior performance of our approach in achieving the dual objectives of preserving identity and facilitating editability. The results highlight the potential and effectiveness of PhotoVerse as a promising solution for personalized text-to-image generation, addressing the limitations of existing methods and paving the way for enhanced user experiences in this domain.

\bibliography{aaai24}
% \bibliography{ref}

\clearpage
\noindent {\LARGE \textbf{Supplementary Material}}
\section{A More Qualitative Results}
\begin{figure}[!ht]
	\includegraphics[width=1.0\linewidth]{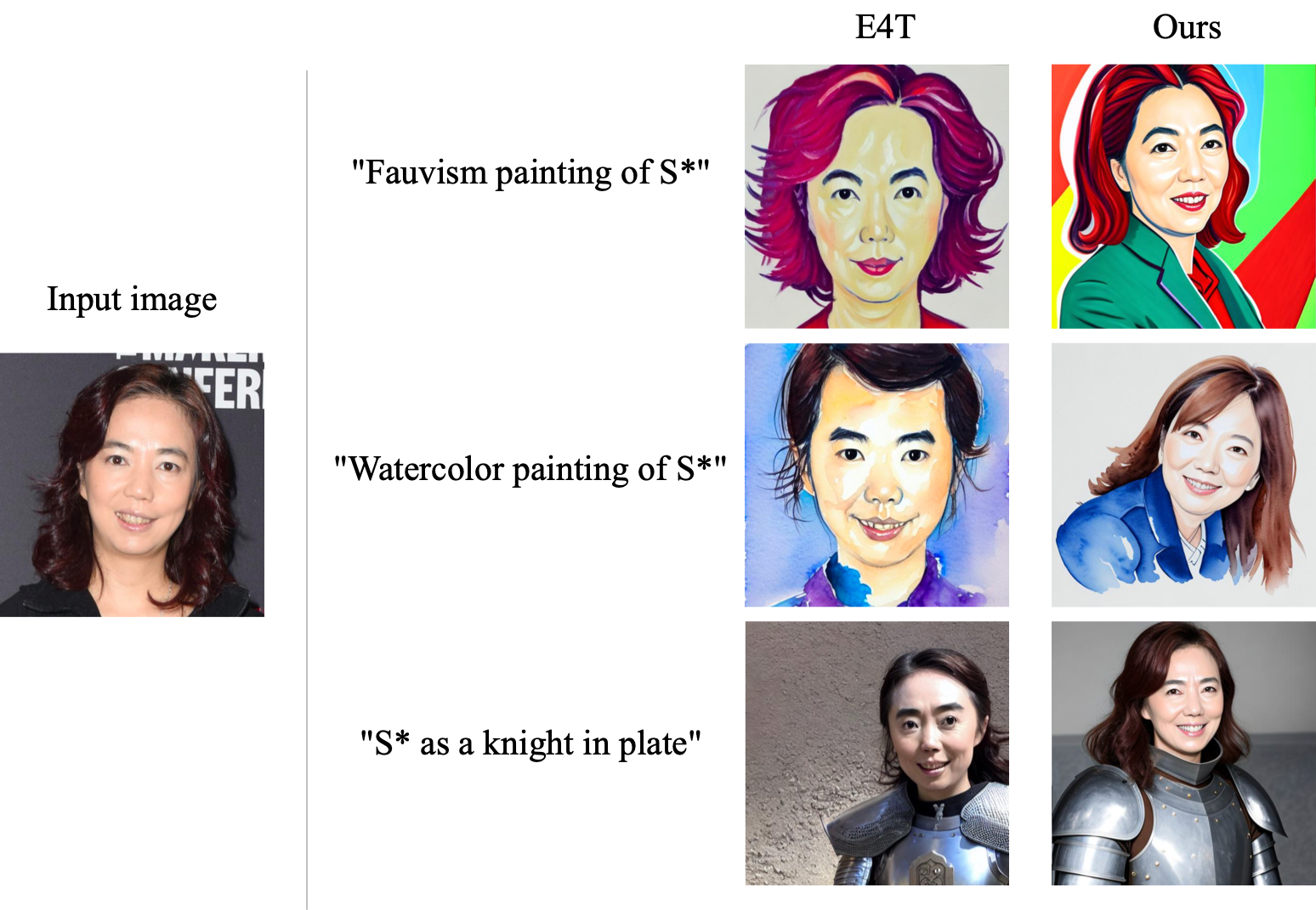} 
	\caption{Comparison results with E4T.}
\label{fig:E4Tcompare} 
\end{figure}

\begin{figure*}[ht]
\centering
	\includegraphics[width=0.5\linewidth]{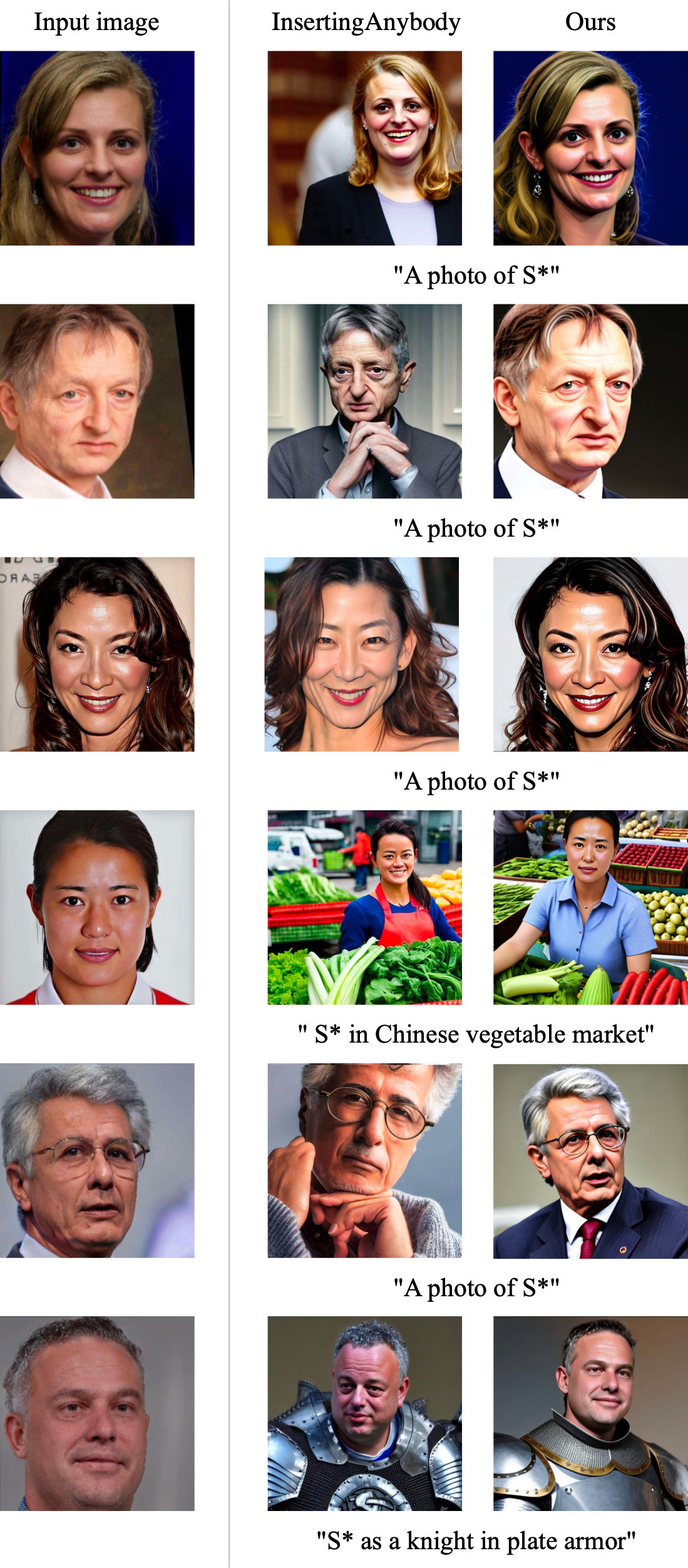} 
	\caption{Comparison results with InsertingAnybody.}
\label{fig:InsertingAnybody} 
\end{figure*}

\begin{figure*}[t]
\centering
	\includegraphics[width=1\linewidth]{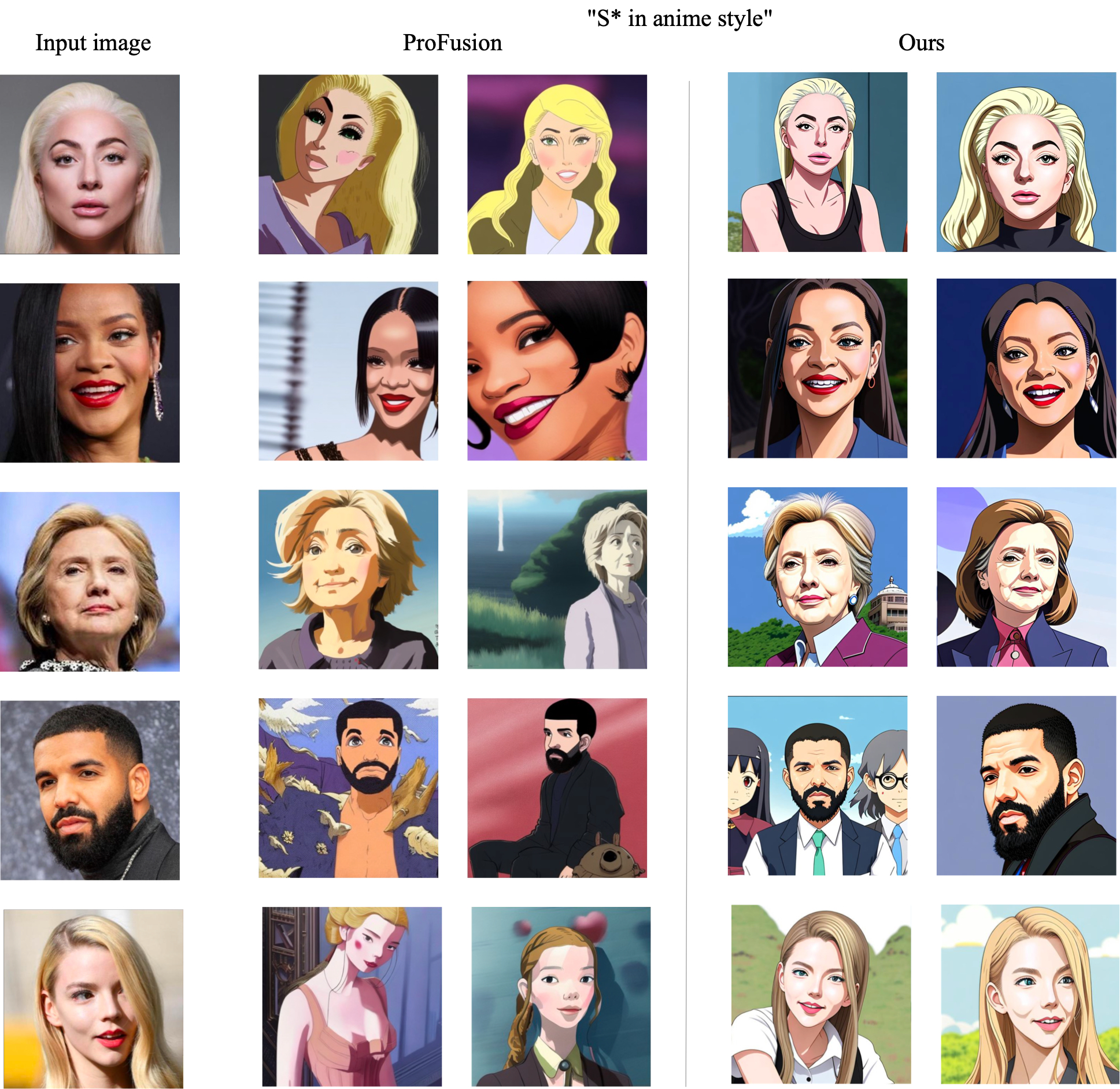} 
	\caption{Comparison results with ProFusion.}
\label{fig:profusion} 
\end{figure*}

\begin{figure*}[t]
\centering
	\includegraphics[width=0.88\linewidth]{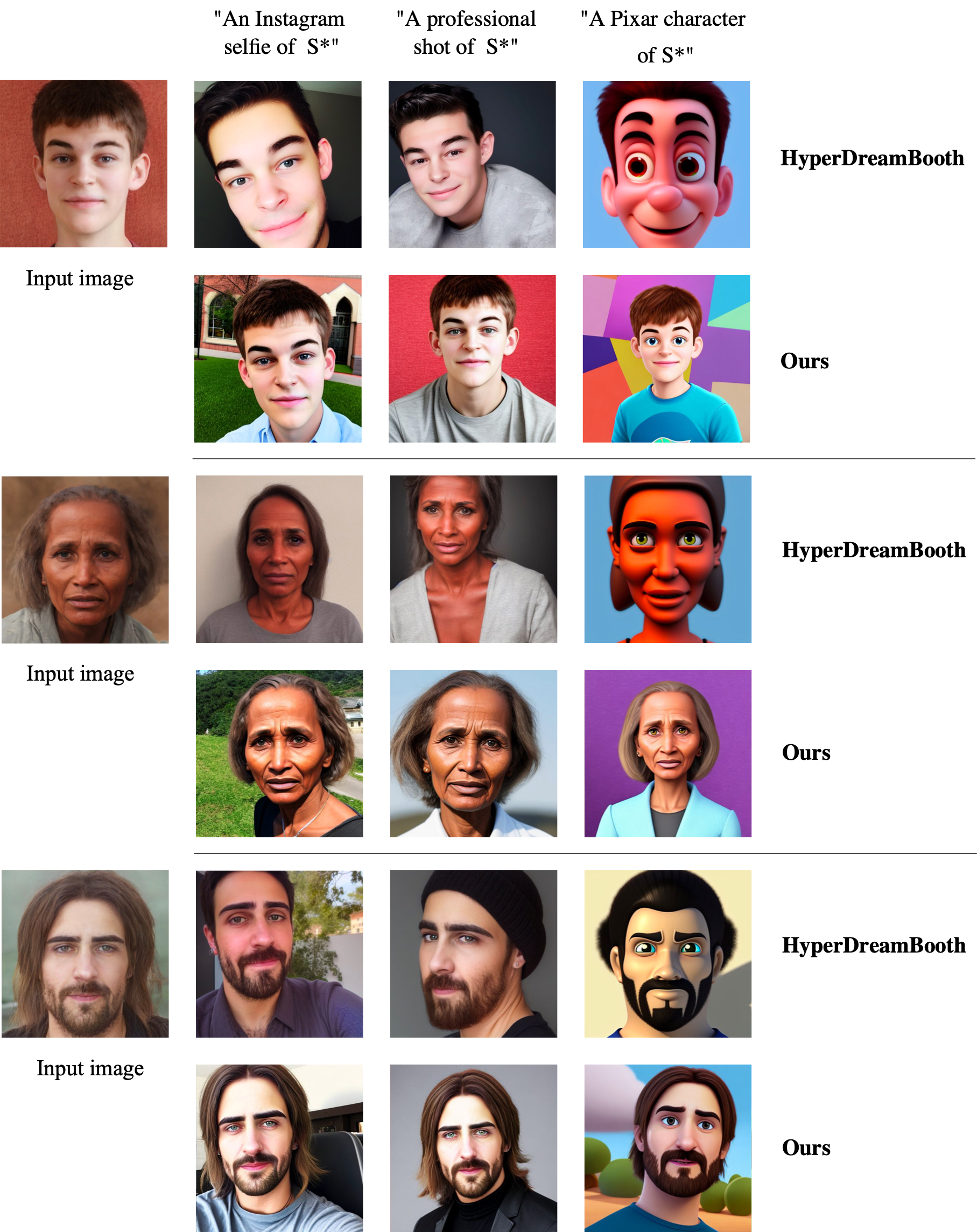} 
	\caption{Comparison results with HyperDreamBooth.}
\label{fig:HyperDreamBooth} 
\end{figure*}

\begin{figure*}[ht]
\centering
	\includegraphics[width=0.8\linewidth]{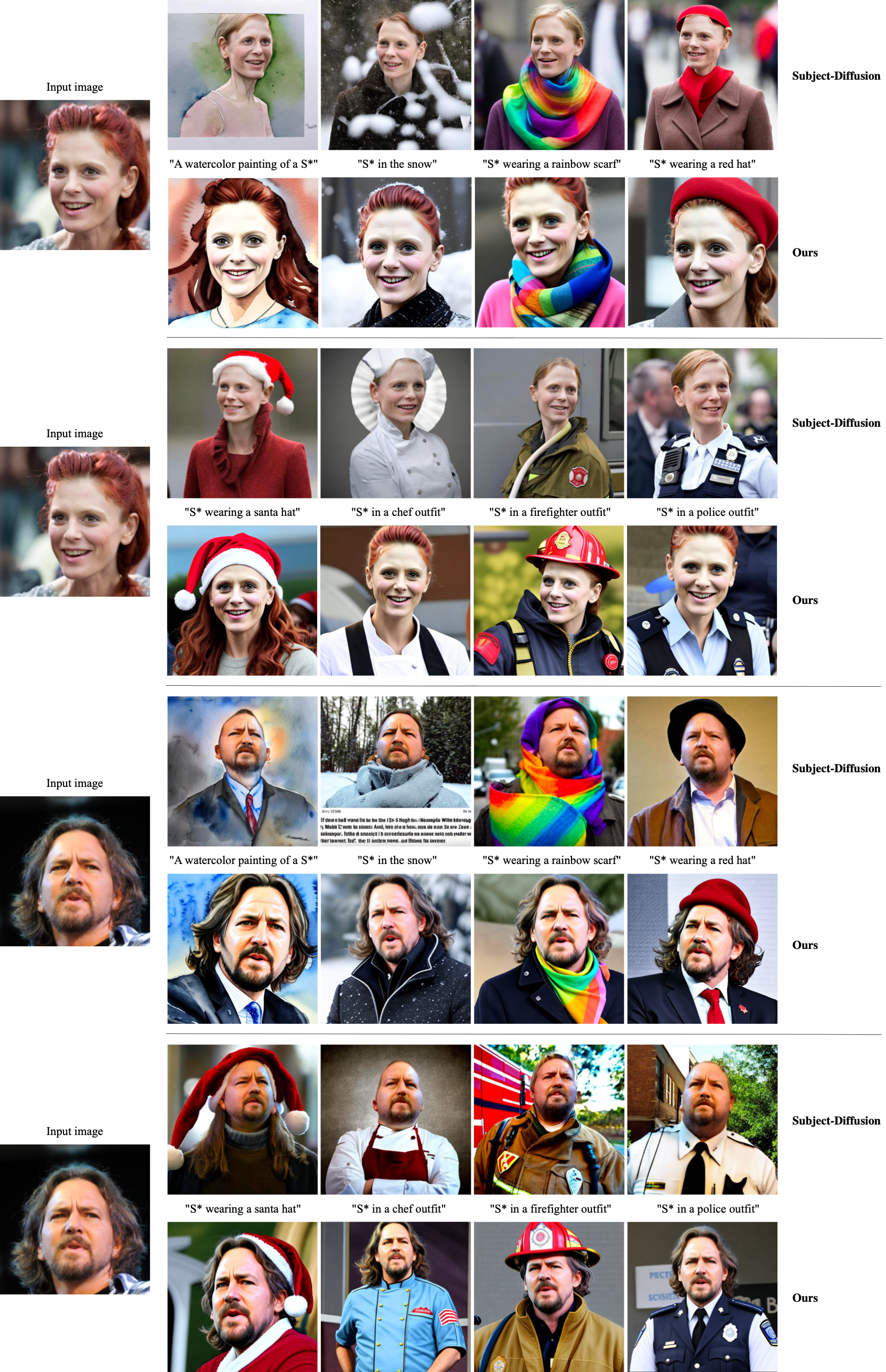} 
	\caption{Comparison results with Subject-Diffusion.}
\label{fig:com-oppo1} 
\end{figure*}

\begin{figure*}[t]
\centering
	\includegraphics[width=0.8\linewidth]{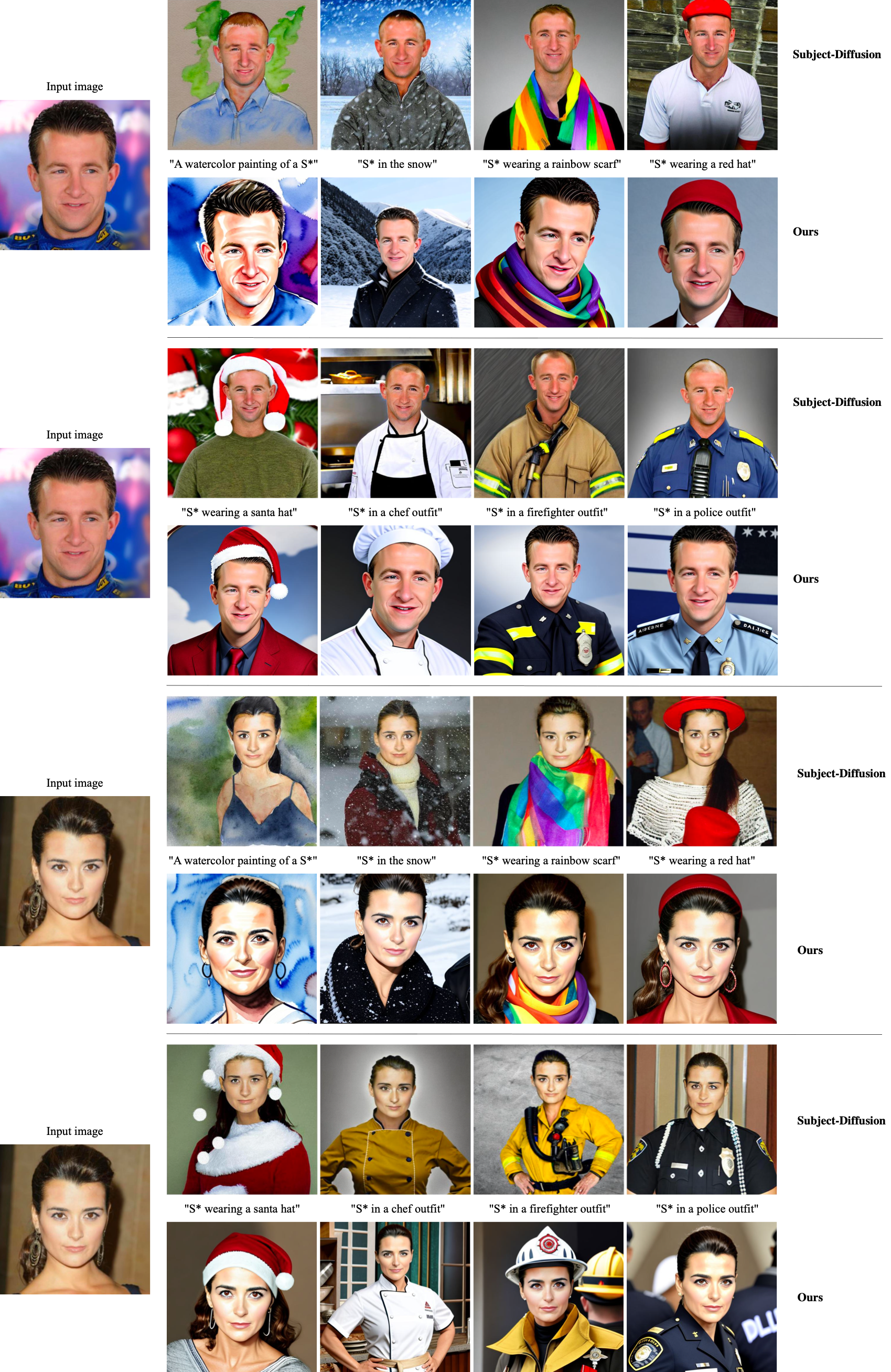} 
	\caption{Comparison results with Subject-Diffusion.}
\label{fig:com-oppo2} 
\end{figure*}

\begin{figure*}[ht]
\centering
	\includegraphics[width=0.85\linewidth]{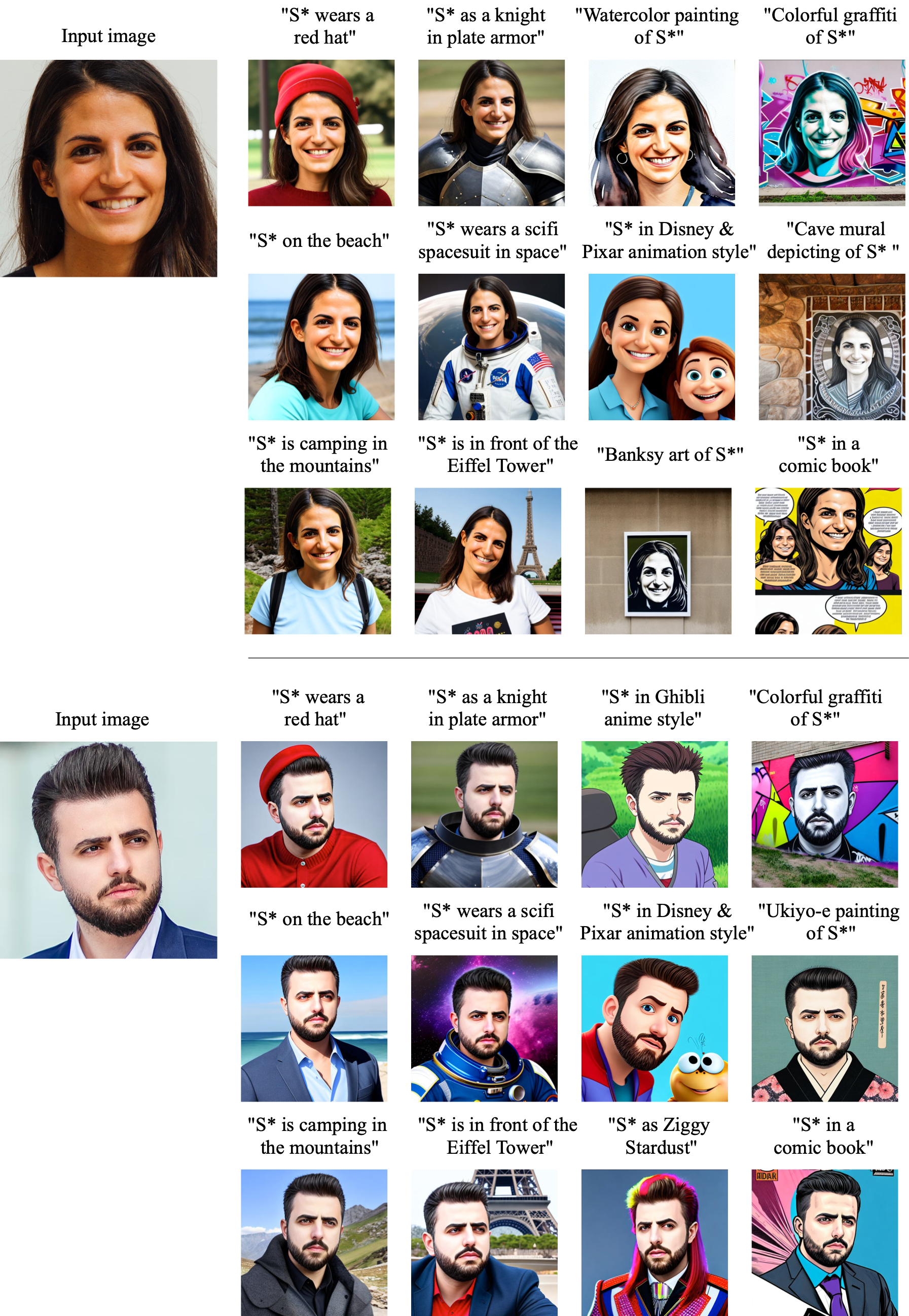} 
	\caption{Results of our PhotoVerse.}
\label{fig:sup-res1} 
\end{figure*}

\begin{figure*}[ht]
\centering
	\includegraphics[width=0.85\linewidth]{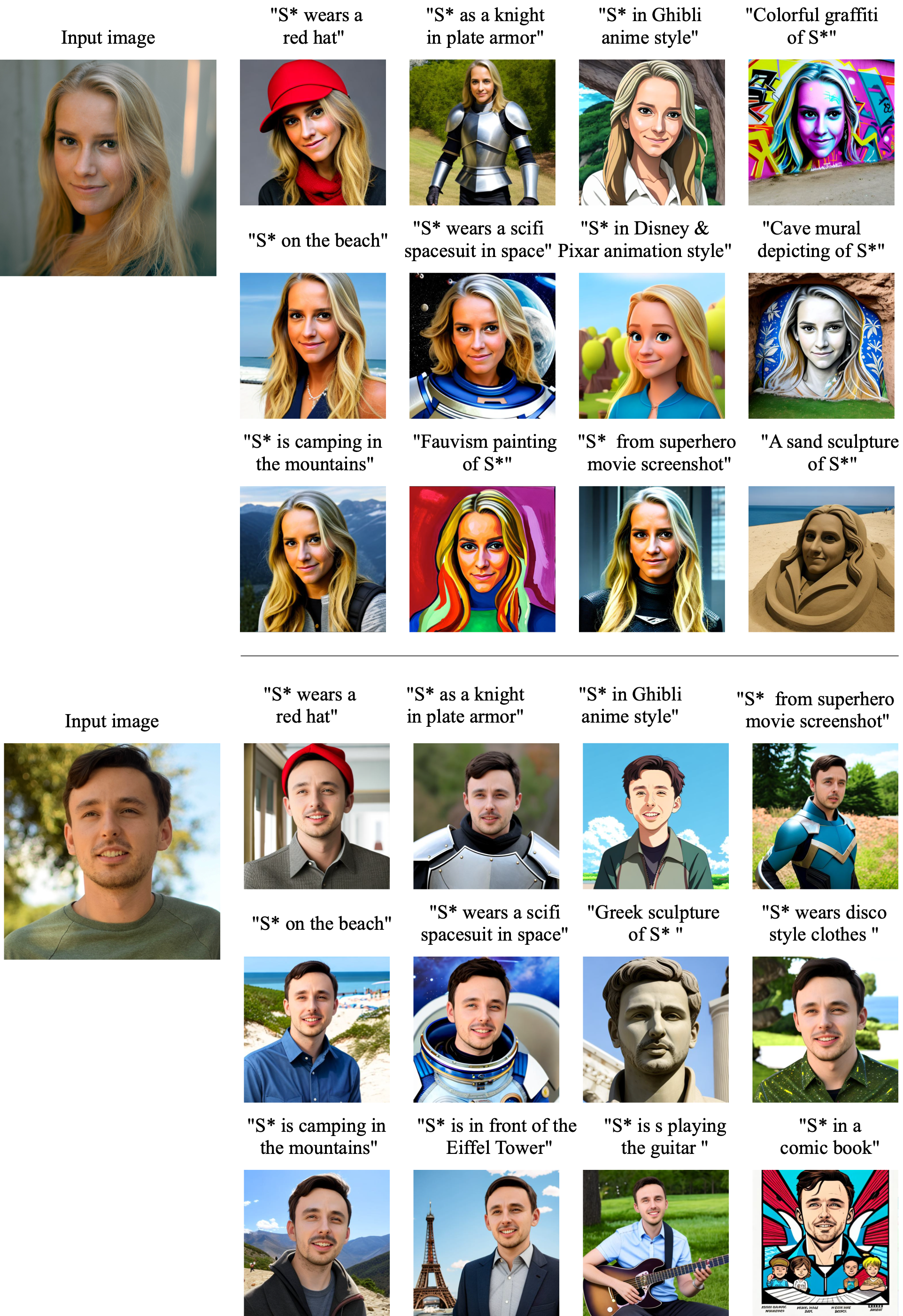} 
	\caption{Results of our PhotoVerse.}
\label{fig:sup-res2} 
\end{figure*}

\begin{figure*}[ht]
\centering
	\includegraphics[width=0.85\linewidth]{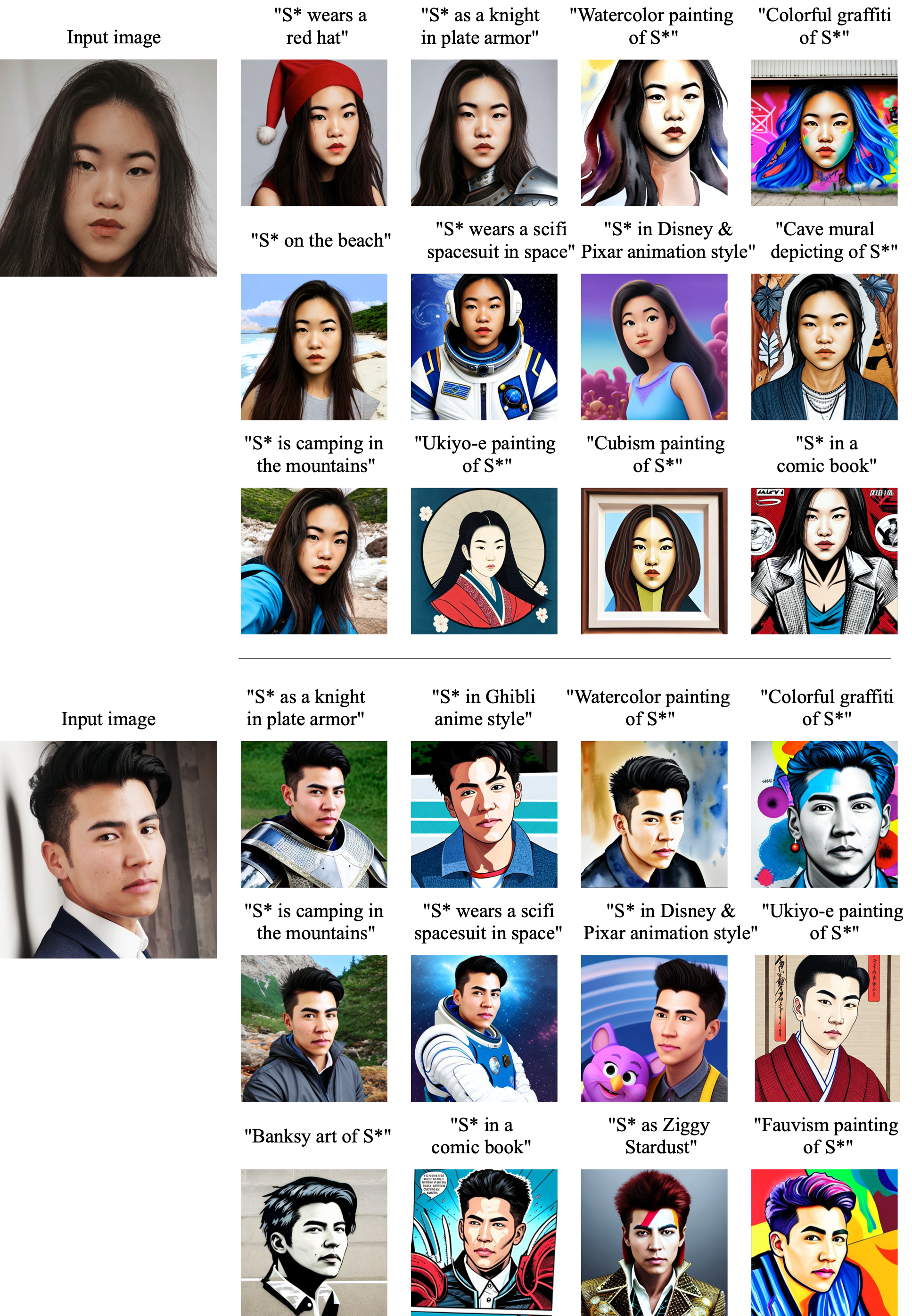} 
	\caption{Results of our PhotoVerse.}
\label{fig:sup-res3} 
\end{figure*}

In Figure \ref{fig:E4Tcompare}, \ref{fig:InsertingAnybody}, \ref{fig:profusion}, \ref{fig:HyperDreamBooth}, \ref{fig:com-oppo1} and \ref{fig:com-oppo2}, we present supplemental comparison results involving state-of-the-art methods E4T \cite{E4T}, InsertingAnybody \cite{yuan2023inserting}, Profusion \cite{zhou2023enhancing}, HyperDreamBooth \cite{ruiz2023hyperdreambooth} and Subject-Diffusion \cite{ma2023subject}, respectively. All the results are taken from corresponding papers. These results show the exceptional performance of our PhotoVerse in terms of fidelity, editability, and image quality. Notably, our approach stands apart from methods E4T \cite{E4T}, InsertingAnybody \cite{yuan2023inserting}, Profusion \cite{zhou2023enhancing} and HyperDreamBooth \cite{ruiz2023hyperdreambooth} by eliminating the need for test-time tuning, while maintaining the ability to generate a single image in $\sim$ 5 seconds.
We also provide more results of our PhotoVerse in Figure \ref{fig:sup-res1}, \ref{fig:sup-res2} and \ref{fig:sup-res3}.

\begin{figure}[!t]
\centering
	\includegraphics[width=0.65\linewidth]{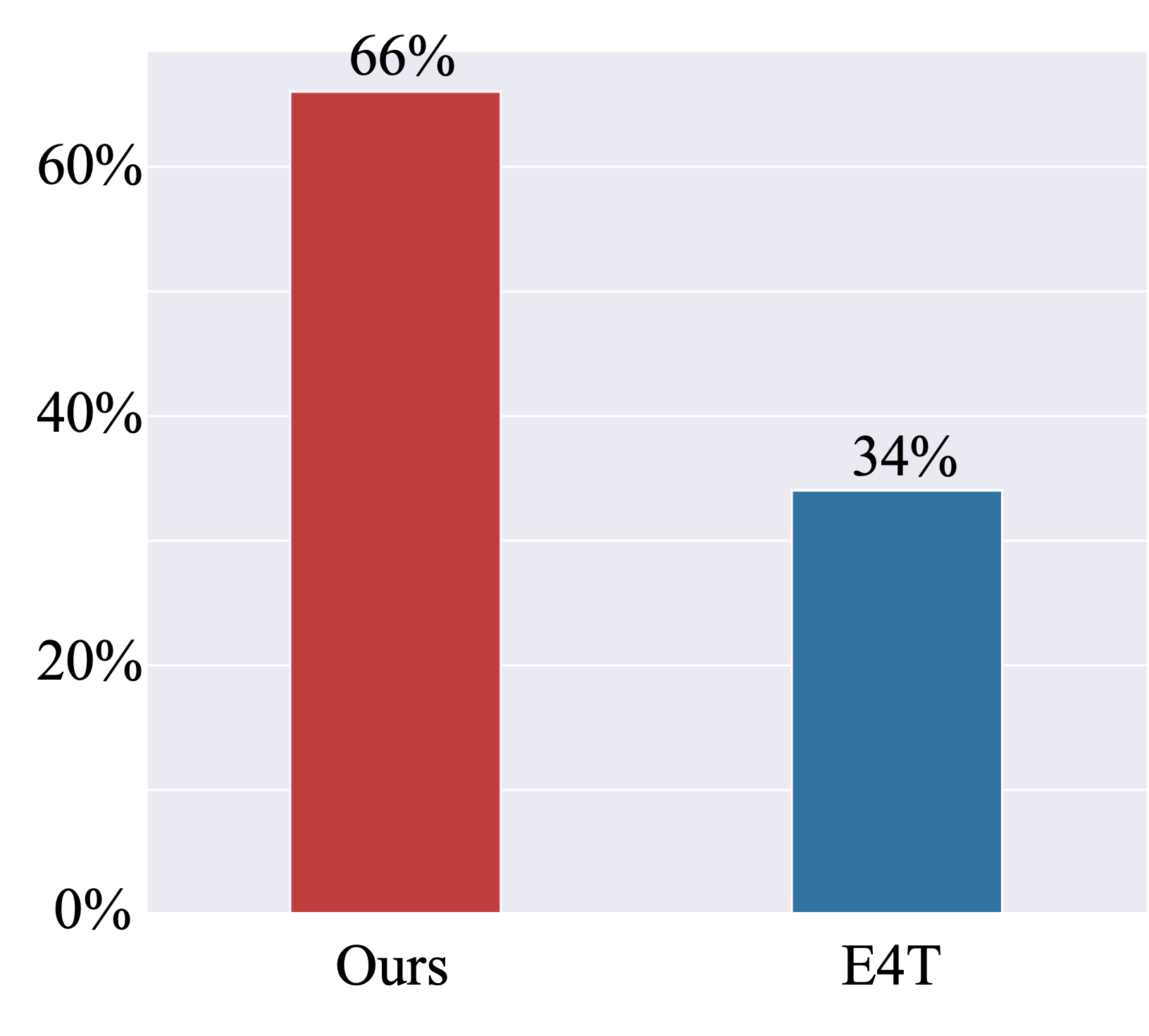} 
	\caption{Results of user study on E4T and our PhotoVerse.}
\label{e4tus} 
\end{figure}

\begin{figure}[!ht]
\centering
	\includegraphics[width=0.65\linewidth]{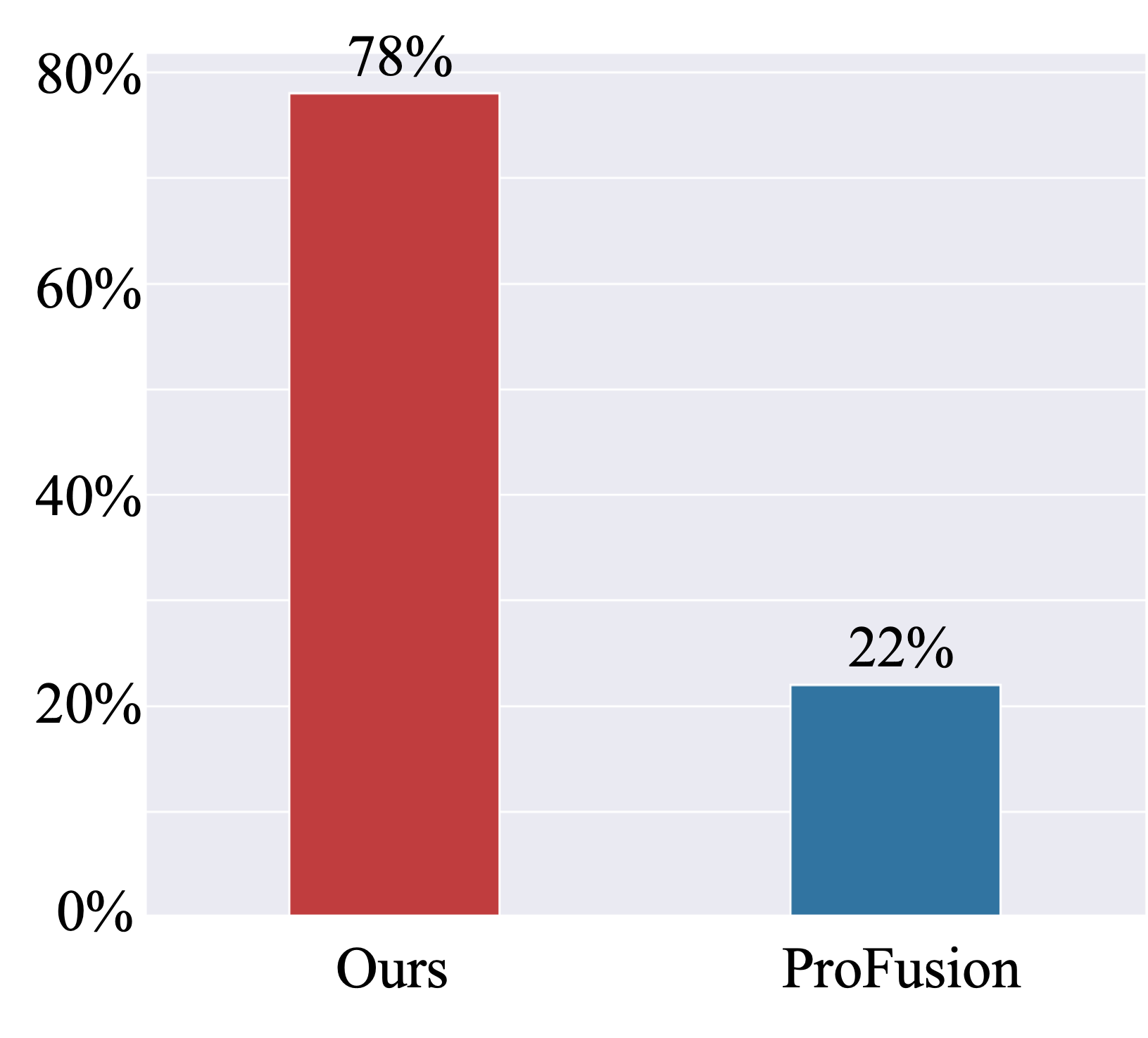} 
	\caption{Results of user study on Profusion and our PhotoVerse.}
\label{profusionus} 
\end{figure}

\section{B Experiment Details}
\subsubsection{Adapter Design}
In our proposed approach, the adapter module is designed to be both straightforward and computationally efficient. It comprises two blocks, with each block consisting of two MLP layers that effectively project the input feature into a 1024-dimensional space. Following this projection, layer normalization and a leaky-ReLU activation layer are applied to further enhance the adapter's performance.

\subsubsection{User Study}
Follow ProFusion \cite{zhou2023enhancing}, we also evaluated E4T \cite{E4T}, ProFusion \cite{zhou2023enhancing}, and our method using human assessment. The results for each method were obtained from the corresponding papers. We designed two separate questionnaires for E4T, ProFusion, and our PhotoVerse for comparison. Each questionnaire consisted of 25 questions. A total of 15 participants took part in the assessment, ensuring a diverse range of perspectives.

\section{C User Study}

As shown in Figure \ref{e4tus} and \ref{profusionus}, we conduct a human evaluation of two recent state-of-the-art methods E4T, Profusion and our PhotoVerse.
For each of the two methods, we requested the participants to provide their preferences among two images generated using different methods, along with the original image and textual instructions. The participants were tasked with indicating their preferred choice according to the subject fidelity and text fidelity.   
According to Figure \ref{e4tus} and \ref{profusionus}, 
our method outperformed the other two recent methods, which are $66\%$ VS $34\%$ and $78\%$ VS $22\%$ respectively.

\end{document}